\pdfoutput=1
\documentclass[11pt]{article}

\usepackage{acl}

\usepackage{times}
\usepackage{latexsym}
\usepackage{multirow}

\usepackage{amsmath}
\usepackage{amssymb}
\usepackage{mathtools}
\usepackage{stfloats}
\usepackage{float}
\usepackage{amsthm}
\usepackage{xcolor}
\usepackage{csquotes}

\usepackage[capitalize,noabbrev]{cleveref}

\theoremstyle{plain}

\theoremstyle{definition}

\theoremstyle{remark}

\usepackage[normalem]{ulem}

\usepackage[T1]{fontenc}
\usepackage[utf8]{inputenc}

\usepackage{microtype}

\usepackage{inconsolata}

\NewDocumentCommand{\zixuan}
{ mO{} }{\textcolor{red}{\textsuperscript{\textit{Zixuan}}\textsf{\small[#1]}}}
\NewDocumentCommand{\pengfei}
{ mO{} }{\textcolor{blue}{\textsuperscript{\textit{Pengfei}}\textsf{\small[#1]}}}
\NewDocumentCommand{\jiaxin}
{ mO{} }{\textcolor{green}{\textsuperscript{\textit{Jiaxin}}\textsf{\small[#1]}}}
\NewDocumentCommand{\heng}
{ mO{} }{\textcolor{red}{\textsuperscript{\textit{Heng}}\textsf{\textbf{\small[#1]}}}}

\title{
Why Does New Knowledge Create Messy Ripple Effects in LLMs? 
}
\author{
Jiaxin Qin$^{1}$\footnotemark[1], Zixuan Zhang$^{1}$, \textbf{Pengfei Yu}$^{1,3}$, Manling Li$^{2}$, \textbf{Heng Ji}$^{1}$\\
  $^1$University of Illinois Urbana-Champaign \\
  $^2$Stanford University\\
  $^3$Boson AI\\
  \texttt{\{qjx0814, zixuan11, pengfei4, hengji\}@illinois.edu} \\
  \texttt{manlingl@stanford.edu}\\
  }

\begin{document}
\maketitle
% \begin{abstract}
% This document is a supplement to the general instructions for *ACL authors. It contains instructions for using the \LaTeX{} style file for EMNLP 2023. 
% The document itself conforms to its own specifications, and is, therefore, an example of what your manuscript should look like.
% These instructions should be used both for papers submitted for review and for final versions of accepted papers.
% \end{abstract}
\footnotetext[1]{Work done during internship at UIUC.}

\begin{abstract}
Extensive previous research has focused on post-training knowledge editing (KE) for language models (LMs) to ensure that knowledge remains accurate and up-to-date.
% One critical criterion to evaluate KE methods is that the knowledge edits are expected to correctly handle \emph{ripple effects}, where edits on one single fact should accurately propagate to its related knowledge.
One desired property and open question in KE is to let edited LMs correctly handle \emph{ripple effects}, where LM is expected to answer its logically related knowledge accurately. In this paper, we answer the question of why most KE methods still create messy ripple effects. 
We conduct extensive analysis and identify a salient indicator, \emph{GradSim}, that effectively reveals when and why updated knowledge ripples in LMs.
GradSim is computed by the cosine similarity between gradients of the original fact and its related knowledge.
% \heng{here should be clearer GradSim is your method name}
We observe a strong positive correlation between ripple effect performance and GradSim across different LMs, KE methods, and evaluation metrics.
Further investigations into three counter-intuitive failure cases (\emph{Negation}, \emph{Over-Ripple}, \emph{Multi-Lingual}) of ripple effects demonstrate that these failures are often associated with very low GradSim. This finding validates that GradSim is an effective indicator of when knowledge ripples in LMs. The code is available.\footnotemark[2]

\footnotetext[1]{https://github.com/JiaxinQin0814/Ripple\_Effect\_Analysis.}

\end{abstract}

\section{Introduction and Related Work}
\label{sec:introduction}

% \heng{in intro give many examples for each counter-intuitive category}

% \chihan{TODO: a teaser figure to explain ripple effect in KE, illustrating why it is important and what a ``successful'' or ``failed'' ripple effect is.}

% Large language models (LLMs) have been widely viewed and used as knowledge bases (KBs) due to their powerful knowledge storage, retrieval, and reasoning abilities in multiple knowledgedomains~\citep{petroni2019language, alkhamissi2022review}. 
% However, real-world knowledge updates and evolves continuously, which motivates 
Large language models (LLMs) can serve as powerful knowledge bases (KBs) thanks to their impressive knowledge storage, retrieval, and reasoning capabilities~\citep{petroni2019language, alkhamissi2022review,zhang2024knowledge}. 
However, real-world knowledge keeps updating and evolving constantly, which motivates extensive research efforts on post-training knowledge editing (KE)~\cite{meng2022locating, memit, yin2023history, zhong-etal-2023-mquake, song2024knowledge, liu2024evedit} to make sure that the knowledge in LMs remains accurate and up-to-date.
% Substantial research efforts~\cite{meng2022locating, DBLP:conf/iclr/MengSABB23, yin2023history, zhong-etal-2023-mquake, song2024knowledge, liu2024evedit} have been made on post-training knowledge editing (KE) and adjustments within LLMs to ensure that the stored knowledge remains current and accurate.
% Constantly pre-training or fine-tuning on them to keep knowledge up-to-date can be resource-consuming. \chihan{citations}
% \zixuan{This is doubtful. Fien-tuning is not really unendurably resource-inefficient.}\chihan{addressed}
% \chihan{citations}.
% \textcolor{orange}{\sout{Yuji: There should be logical connection from the previous efficiency papers to the next introduced ripple effects problem. What's the relationship between efficiency issues to ripple effects? If there is no explicit connection between them, should avoid talking about efficiency to distract the readers.}}
% Another problem arises: nearly no knowledge is isolated from other facts, so adding a single new fact often necessitates numerous consequential knowledge updates.
% \chihan{TODO: why ripple effect is desired}\jiaxin{done}
\begin{figure}[htbp]
    \centering
     \includegraphics[width= 0.49\textwidth]{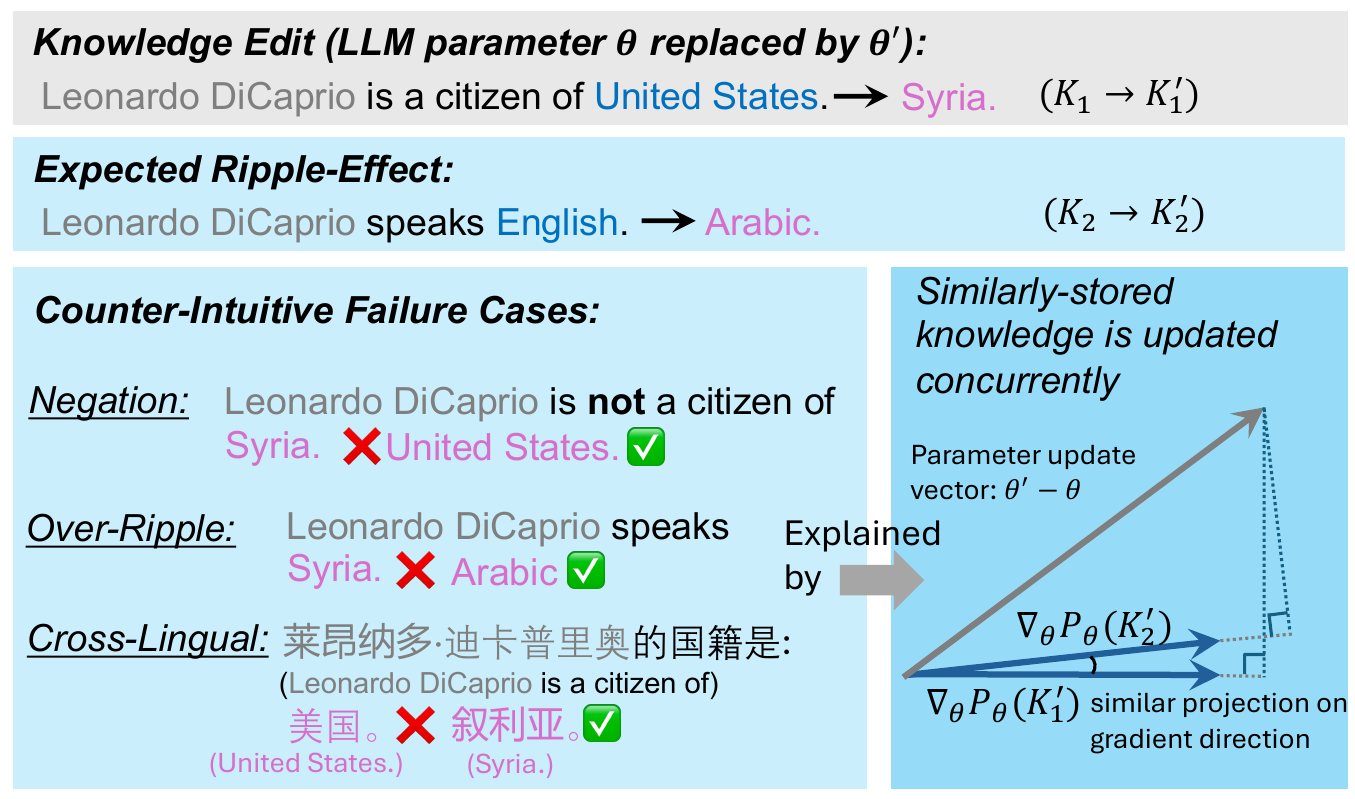}
    \caption{An illustration of ripple effects in LLM knowledge editing. Our work empirically demonstrates the positive correlation between gradient similarity explains a large portion of the ripple effect. Furthermore, messy similarities between knowledge points create several counter-intuitive ripple effect failures.
    % \chihan{can we also draw and show that ``the projection on similar gradients are also similar''?} \heng{should provide Engish translation for the Chinese example}
    }
     % \chihan{``Our work empirically demonstrates the positive correlation between gradient similarity explains a large portion of the ripple effect. Furthermore, messy similarities between knowledge points create several counter-intuitive ripple effect failures.''}
     \vspace{-1cm}  
    \label{fig:illustration}
\end{figure}

Previous research has proposed numerous evaluation metrics to ensure the efficiency and consistency of these editing methods. 
Among them, one critical criterion is the ability of the KE method to handle \emph{ripple effects}~\cite{rippleedit}, where a single edit should automatically and accurately propagate to related facts.
For example, suppose an edit changes \emph{Leonardo DiCaprio}'s nationality to \emph{Syrian}. The model should automatically update its related information, such as knowing that his primarily used language is now \emph{Arabic}.
Such a task is very challenging because it requires the model to correctly understand and infer complex relationships among knowledge elements and accurately locate their parametric storage in order to perform the edits. 
% However, unlike in explicit knowledge bases, prior work reveals insufficient ``ripple effect'' in multiple LLMs combined with various knowledge editing techniques. 
Empirically, even though direct knowledge edits typically achieve over $>90\%$ accuracy, the success rates of ripple effects struggle to exceed $50\%$ across all recent KE methods, even on the simplest task in RippleEdits~\cite{rippleedit}.

In this paper, we answer the intriguing research question of when and why updated knowledge ripples in language models. 
We hypothesize that the knowledge storage among parameters plays a critical role in determining the ripple effects between knowledge facts. 
A messy relationship among knowledge elements can make achieving a successful ripple effect intractable or impossible.
% We aim to identify the key contributory factors and provide an in-depth explanation from the perspective of model parameters and knowledge distributions.
% \chihan{``\textit{We hypothesize that the knowledge storage among parameters plays a critical role in determining the ripple effects between knowledge facts. A messy relationship among knowledge can render a successful ripple effect intractable or impossible.}''}
Intuitively, the similarity of knowledge storage should be an important factor, as 
% related facts can interact and influence each other more easily if their parametric storage locations are similar.
% \chihan{``knowledge represented by similar parameters will respond similarity to parameter updates during knowledge editing.''}
knowledge represented by similar parameters will respond similarly 
% \heng{did you want to say 'respond similarly'?} 
to parameter updates during knowledge editing.
Following this intuition, we conduct extensive analysis and identify a salient indicator that strongly reveals how likely an updated fact 
% \heng{'knowledge' is not countable, so maybe use 'knowledge element' or 'fact' throughout the paper}
will ripple in a language model: the \emph{cosine similarity} between the \emph{gradients of the related knowledge facts} (GradSim).
We use gradients to represent knowledge storage distribution in LMs because they indicate which parameters in the model
% aims to adjust and quantify these adjustments when encountering new knowledge.
are responsible for increasing or decreasing the likelihood of answering certain knowledge.
% \chihan{``are responsible for increasing or decreasing the likelihood of answering certain knowledge.''}
% dominantly contributes to the success and failure of knowledge ripples: the similarity of \emph{knowledge storage distribution} between the original and rippled knowledge facts.
% Such a finding aligns with our intuition, as it would be much easier for related facts to interact and influence each other if their parametric locations are similar.
% To provide a more rigorous mathematical framework, we use gradients as indicators of the parametric storage distribution and employ cosine distance to describe their similarities.
We observe a strong positive correlation between ripple effect performance and the cosine similarity of gradients across different LMs, editing methods, and evaluation metrics, with a Pearson correlation metric reaching as high as 0.85.
% \begin{figure}[htbp]
%     \centering
%      \includegraphics[width= 0.49\textwidth]{figures/illustration_ke.pdf}
%     \caption{An example of ripple query after knowledge editing.}
%     \label{fig:illustration}
% \end{figure}

The hypothesis and analysis above predict a counter-intuitive phenomenon: knowledge with similar parameter-storing locations, even if logically unrelated or contradictory, will create positive ripple effects toward each other, and vise versa.
% \chihan{``The hypothesis and analysis above predicts a counter-intuitive phenomenon: \textit{knowledge with similar storing parameters, even if logically unrelated or contraditory, will create positive ripple effects toward each other, and vise versa.}''}
Viewing GradSim as an indicator of ripple effects, we verify this paradox above by discovering and explaining three specific ripple effect cases: \textbf{\emph{Negation}}, \textbf{\emph{Over-Ripple Errors}} and \textbf{\emph{Cross-Lingual Transfer}}.
As illustrated in Figure~\ref{fig:illustration}, assuming that a knowledge edit changes the citizenship of \emph{Leonardo DiCaprio} from \emph{United States} to \emph{Syria}. 
In \textbf{\emph{Negation}}, LMs with different sizes including GPT2-XL and LLaMA-2 unexpectedly answer the negated query \emph{Leonardo is not a citizen of} still by \emph{Syria} instead of logically correct answers such as \emph{United States}.
Moreover, in \textbf{\emph{Over-Ripple Errors}}, the LMs over-memorize the edit target \emph{Syria}, and tends to always answer \emph{Syria} even when asked about other topics such as language.
In \textbf{\emph{Cross-Lingual Transfer}}, even the most powerful cross-lingual LMs could make mistakes when asked about the edited knowledge in a different language.
All of these ripple-effect cases are commonly encountered in real-world applications, but are more challenging and often experience counter-intuitive failures with current LMs and KE methods.
In our experiments, we demonstrate that the model's failure in these cases is strongly correlated with a too small GradSim, the similarity in knowledge distributions within LMs.

\section{GradSim: A Ripple Effect Indicator}
In this section, we formally introduce GradSim, a ripple-effect indicator based on the knowledge storage similarity between related knowledge.
We use $x$ and $y$ to denote a pair of original fact and its related knowledge respectively, and we use $(q_x, a_x)$ and $(q_y, a_y)$ to represent query-answer pairs based on the corresponding knowledge facts.
For instance, if $q_x$ and $a_x$ are \emph{<Leonardo DiCaprio is a citizen of>} and \emph{<United States>} respectively, then one example pair of $q_y$ and $a_y$ could be \emph{<Leonardo DiCaprio speaks>} and \emph{<English>}.
Given a query $q_x$, typical KE methods update $a_x$ to a new answer $a'_x$ by applying an update on the model parameters~$\theta$, and ripple effect evaluations expect that the LM can automatically find the correct $a'_y$ when asked $q_y$.
Based on our hypothesis, knowledge represented by similar parameteres will respond similarly to parameter updates during knowledge editing.
We employ the gradient to model the storage of knowledge within an LLM, and use the cosine similarity to measure the proximity between the storage distribution of two pieces of knowledge:

\begin{equation*}
\begin{aligned}
\textit{GradSim}(x,y) &= \\ \cos(\nabla_\theta P_\theta(a_x | q_x), 
&\nabla_\theta P_\theta(a_y | q_y))
\end{aligned}
\end{equation*}

% \begin{align*}
    
% &\textit{GradSim}(x,y) =& \\
% &\cos(\nabla_\theta P_\theta(a_x | q_x)\nabla_\theta P_\theta(a_y | q_y))&
% \end{align*}

% \paragraph{Discussion:} \chihan{redundant now} We claim that this metric also reflects how similarly the two knowledge points are stored within LLMs parameters. Intuitively, the gradient $\nabla_\theta P_\theta(a_i | q_i)$ indicates what parameters are ``responsible'' for maximizing the probability of predicting $a_i$ on query $q_i$. If the parameters $\theta$ are updated according to this direction, $a_i$ is more likely to be generated after $q_i$, and vise versa. Then the cosine similarity $\cos()$ between two knowledge points' gradients tells us the overall level of overlap between their ``responsible parameters''.

\section{Experiments}
We assess the effectiveness of GradSim by empirically examining its correlation with ripple effect performance, aiming to determine if it reliably indicates successful knowledge propagation in language models. 
Furthermore, we analyze three typical counter-intuitive failure cases in detail to understand the role of GradSim in these situations.

\begin{figure*}[ht!]
  \centering
  \includegraphics[width=\textwidth]{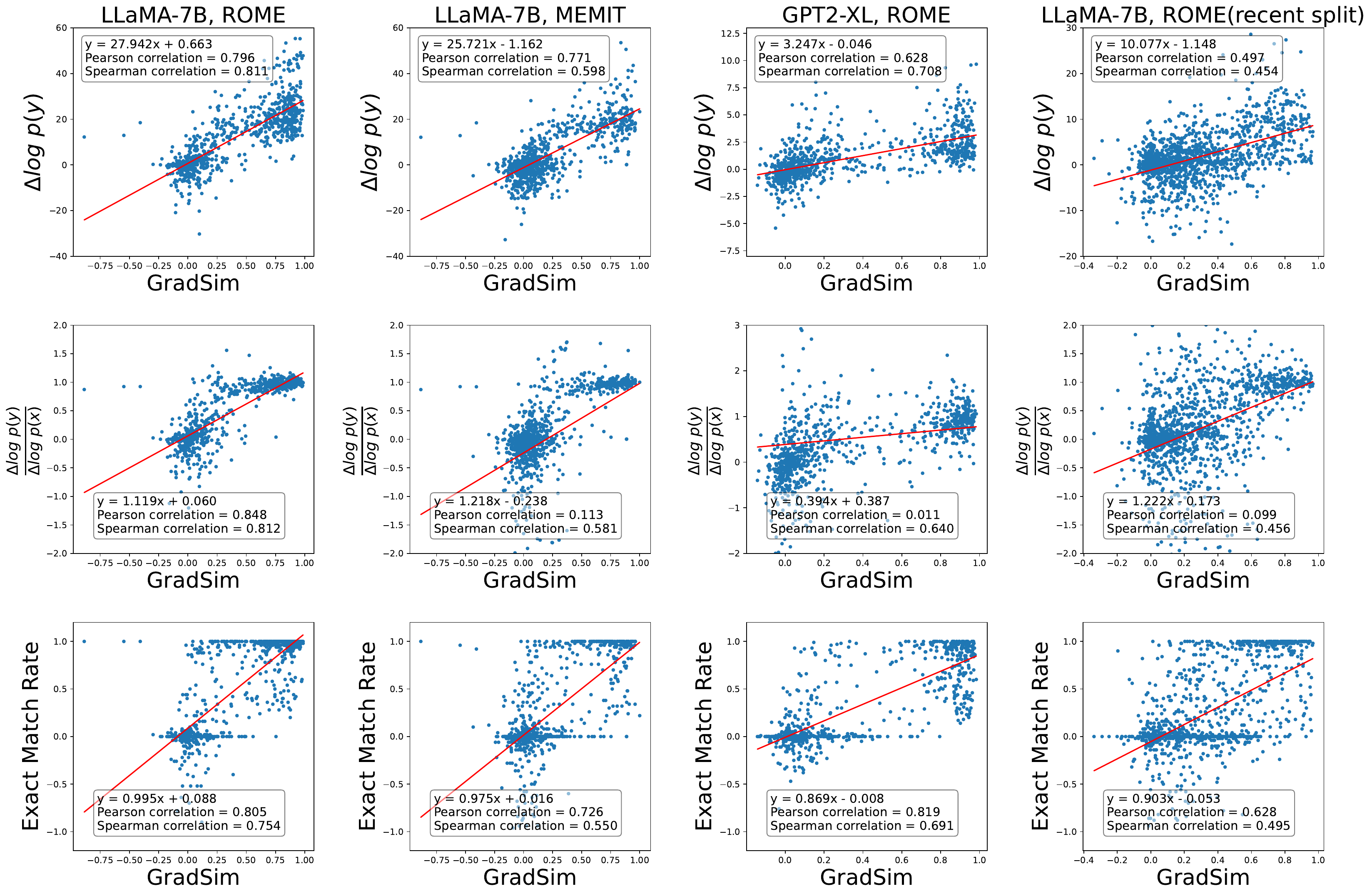} 
  \caption{Main results of evaluating the correlation between ripple effect performances and GradSim values.}
  \vspace{-0.4cm}
  \label{fig:wideimage}
\end{figure*}
\subsection{Data, Models, and KE Methods}
In our experiments, we mainly employ \emph{RippleEdits} \cite{rippleedit}, the most widely-used benchmark for evaluating ripple effects for knowledge editing methods in LLMs.
We mainly use the \emph{popular} split in \emph{RippleEdits} because popular entities are more likely to be recognized by language models. This approach helps to minimize any side effects resulting from the model's lack of knowledge, allowing us to focus on its reasoning abilities.
To ensure a comprehensive evaluation, we also demonstrate results in the \emph{recent} split as shown in Figure~\ref{fig:wideimage}.
In our experiments, we consider two typical knowledge editing (KE) methods: ROME~\cite{meng2022locating} and MEMIT~\cite{memit}, which are based on the locate-and-edit approach to modify model parameters. 
For language models, we evaluate both larger models like LLaMA2-7B~\cite{llama2} and smaller models like GPT-2 XL~\cite{gpt2} to ensure a comprehensive evaluation and maintain consistency with the original settings in ROME and MEMIT.

% to examine the correlation between the internal indicator and our approach to modeling knowledge similarity. 
% TRippleEdits offers a structured framework with six specific evaluation criteria designed to assess the impact of factual edits beyond the scope of the editing action itself. These criteria aim to determine whether changes should be maintained or revised based on their subsequent effects. 
% The dataset comprises about 5,000 entries, each featuring a factual edit accompanied by a series of test queries. These queries are instrumental in evaluating the success of each edit, focusing on its ``ripple effect''—the extent to which the edit influences related facts within the dataset. 

\subsection{Evaluation Metrics of Ripple Effects}
% \chihan{``As likelihood values can be very small as the sequence length increases, we use log-likelihood score $\log P(a_i | q_i)$ as a more sensitive score.''}
% $$\text{NLL}_{\theta}(a_i|q_i) =  -\log P_{\theta}(a_i|q_i)$$
To ensure the validity of our experiment results, we consider multiple evaluation metrics assessing how well the model performs in answering ripple-effect queries.
These metrics include both accuracy-based measures, such as the exact match rate, and more quantified likelihood metrics, such as the absolute and relative gains in likelihood.
\paragraph{Exact-Match (EM) Rate}
Similar to \cite{rippleedit}, we first consider accuracy-based metrics to calculate the proportion of correct answers the model generates from multiple random sampling choices.
For each ripple query $q_y$, we sample 50 generated answers with a temperature 0.7, and compute the proportion of answers that include the correct answer.
Our metric differs slightly from that in ~\cite{rippleedit}, as we need to compute performance for each individual data point to analyze the overall correlation.
The maximum length for generation is set to a small size of 15, as we believe that the answer is expected to appear early in a cloze-test query format.

\paragraph{Absolute Likelihood Gain}
We also examine the answer probabilities to obtain a more detailed and quantifiable assessment of the performance.
As probability values could be very small as the sequence length increases, we use log-likelihood score $\log P(a'_y | q_y)$, and measure its absolute gain on the correct answers before and after editing:
\begin{equation*}
    \Delta \log P(y)= \log P_{\theta'}(a'_y \mid q_y) - \log P_{\theta}(a'_y \mid q_y).
\end{equation*}
\paragraph{Relative Likelihood Gain}
% To account for the inherent difficulty of the knowledge editing process, we also consider a relative likelihood gain. 
This metric is formulated by dividing the absolute gain of ripple effects by the absolute gain of the original fact, thereby normalizing the difficulty of the knowledge editing itself.
\begin{equation*}
    \frac{\Delta \log P(y)}{\Delta \log P(x)} = \frac{\log P_{\theta'}(a'_y \mid q_y) - \log P_{\theta}(a'_y \mid q_y)}{\log P_{\theta'}(a'_x \mid q_x) - \log P_{\theta}(a'_x \mid q_x)}.
\end{equation*}

\subsection{Main Results}
We conduct a comprehensive correlation analysis across different language models, knowledge editing methods, and performance metrics for ripple effects, with the results illustrated in Figure~\ref{fig:wideimage}. 
A strong positive correlation is observed between ripple-effect performances and the GradSim values, validating that gradient-based knowledge storage similarity is a reliable indicator of ripple effects. 
Additionally, we observe the emergence of two distinct clusters in the figure. 
This clustering likely occurs because the data points can be categorized into successful and unsuccessful edits. 
Successful edits result in a significant improvement on performance, placing them in the upper successful cluster, while unsuccessful edits tends to remain in the unsuccessful cluster. 
% Despite this separation, we still observe a strong positive correlation overall.

% \begin{enumerate}
%     \item main result (cosine and NLL)
% % \begin{table*}[htbp]
% %   \centering
% %   \caption{Main plot.}
% %   \label{tab:table1}
% %   \begin{tabular}{c|c|c|c|c|c|c|c}
% %     \hline
% %     model& edit method& slope & $p_{slope}$ & spearman & $p_{spearman}$ & pearson &$p_{pearson}$\\
% %     \hline
% %     gpt2-xl & ROME & & &  & & & \\
% %     gpt2-xl & MEMIT & & &  & & & \\
% %     % gpt2-xl & fine-tune  & & &  & & & \\
% %     llama-7b & ROME & & &  & & & \\
% %     % llama-7b & MEMIT & & &  & & & \\
% %     % llama-7b & fine-tune  & & &  & & & \\
% %     \hline
% %   \end{tabular}
% % \end{table*}

%     \item split the relation according to conditional query/edited query(choose some of the results to plot)
%     % \item \ {test other 2 knowledge neurons identification methods}
%     \item test other sub-datasets in RippleEdits(to many)
% \end{enumerate}

\subsection{Counter-Intuitive Failure Cases}
\begin{figure}[htbp]
  \centering
  \includegraphics[width=\linewidth]{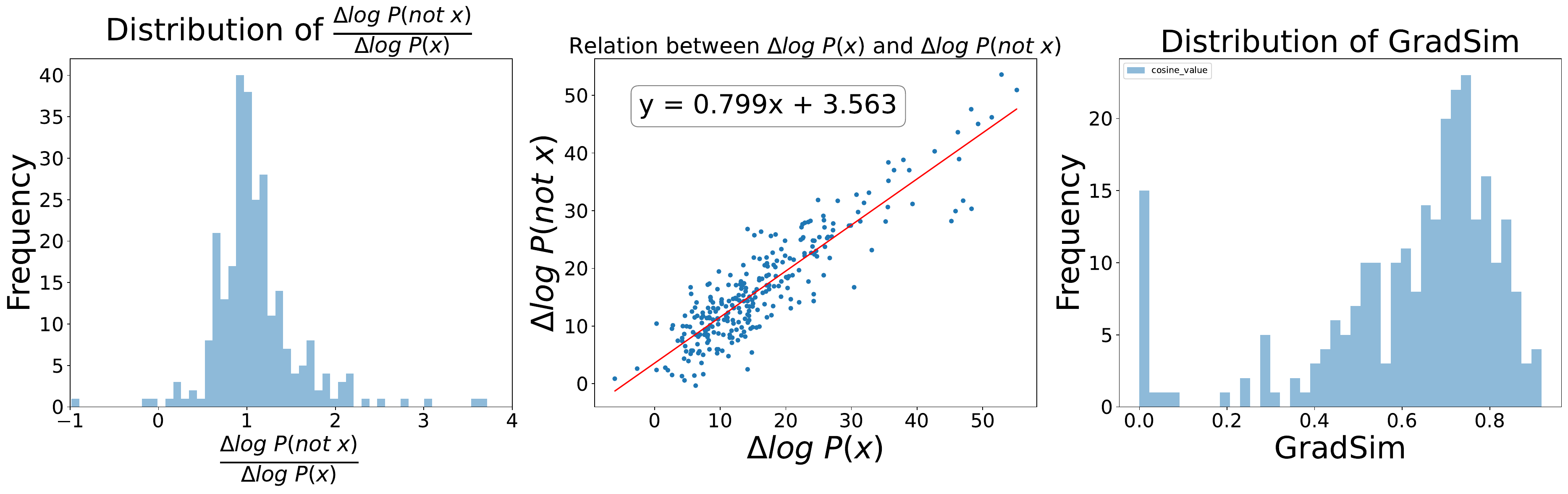}
  \caption{Correlation between original and negated facts on likelihood change, and the distributions of GradSim and likelihood ratios between original and negated facts. }
  \vspace{-0.55cm}
  \label{fig:Negation}
\end{figure}

\paragraph{Negation} Negation is one of the most straight-forward ripple effects where the model is expected to answer a negated query after an editing is applied.
For example, after editing the nationality of \emph{Leonardo DiCaprio} as \emph{Syrian}, the model should be able to avoid \emph{Syria} given a negated query like ``\emph{Leonardo DiCaprio is \textbf{not} a citizen of}''.
However, both smaller-sized LMs like GPT-2 and larger-sized LMs like LLaMA still answers ``Syria'' to this query and simply ignore the negation inside the sentence.
In Figure~\ref{fig:Negation}, we first visualize both the values and gains of model likelihoods for the original and negated facts. 
The results demonstrate a strong positive (almost linear) correlation, indicating a severe problem of negation failures.
In terms of GradSim values, we find that the gradient similarities between the original and negated facts are very high, suggesting that the original and negated facts are entangled in similar knowledge storage locations.

\begin{figure}[htbp]
  \centering
  \includegraphics[width=\linewidth]{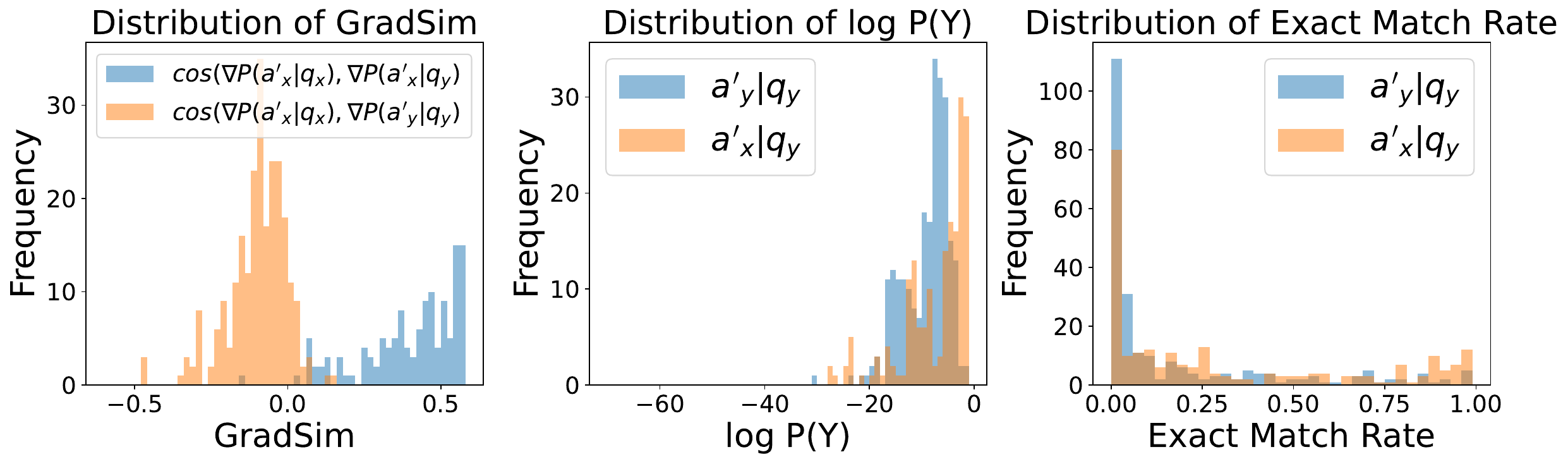}
  \caption{The distributions of GradSim values and ripple effect performances.}
  \label{fig:over-ripple}
  \vspace{-0.5cm}
\end{figure}

\paragraph{Over-Ripple Errors}
The over-ripple problem refers to the situation where, after a knowledge edit, the LM only memorizes the edited target itself and continues to provide this target as the answer even when asked about other knowledge that is related.
For example, after editing the nationality of \emph{Leonardo DiCaprio} as \emph{Syrian}, the model will still answer \emph{Syria} even when asked about the primary language Leonardo is speaking (the correct answer should be \emph{Arabic}).
In Figure~\ref{fig:over-ripple}, we first visualize GradSim distributions on $(q_y, a'_x)$ and $(q_y, a'_y)$ respectively, and we can observe that the edited target $a'_x$ (e.g., \emph{Syria}) has a much higher gradient similarity compared to the correct answer $a'_y$ (e.g., \emph{Arabic}).
This explains the similar performance of answering both the correct and incorrect answers and indicates the occurrence of over-ripple errors.

\begin{figure}[htbp]
  \centering
  \includegraphics[width=\linewidth]{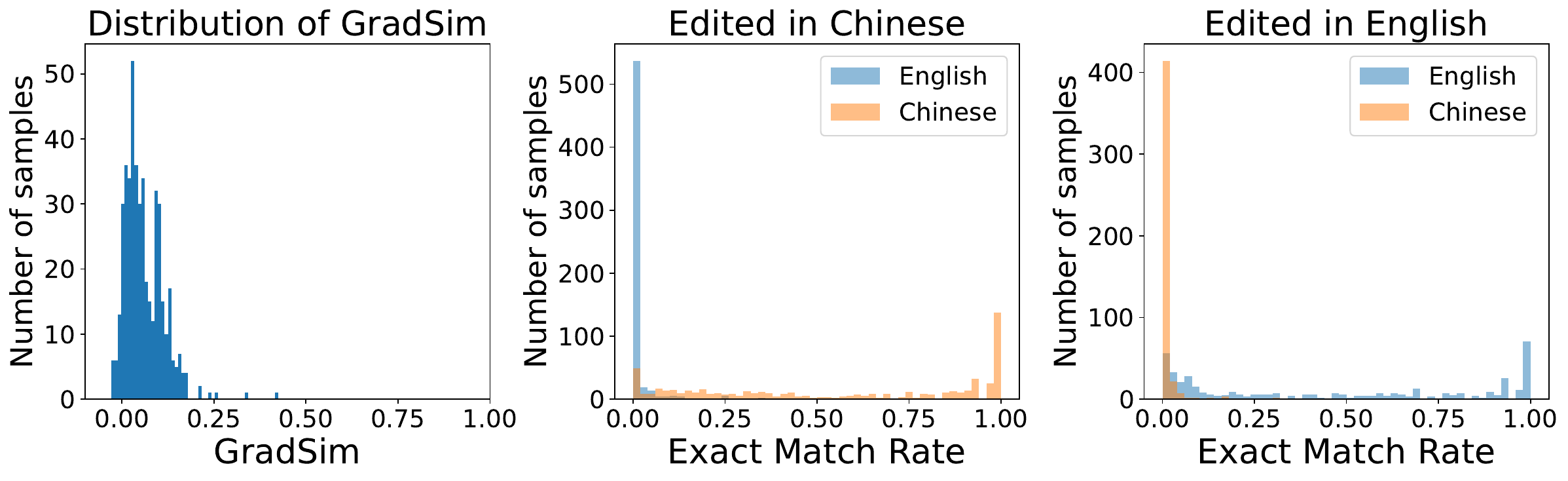}
  \caption{GradSim and performance distributions when editing on one language and testing on another.}
  \label{fig:crosslingual}
  \vspace{-0.55cm}
\end{figure}

\paragraph{Cross-Lingual Transfer}
The problem of cross-lingual transfer is defined as the ability to edit a piece of knowledge in one language and have the model still provide the correct answer when asked a question in another language. 
% This kind of ripple effect is common in real-world scenarios and relatively easy for humans, but it is counter-intuitively challenging for LMs.
We study the role of GradSim by visualizing the distribution of GradSim values and the ripple effect performance across different languages.
We employ Baichuan~\cite{yang2023baichuan}, the state-of-the-art bilingual model for Chinese and English. 
As shown in Figure~\ref{fig:crosslingual}, while the performance on the target language remains low, the GradSim values are also very low, primarily distributed near zero.
GradSim works as a reliable indicator in this special ripple-effect case for cross-lingual transfer.

\section{Conclusion}
Through extensive experiments and analysis, we propose GradSim, computed as the cosine similarity between gradients of the original fact and its related knowledge, as a crucial indicator for the effectiveness of the ripple effects. 
The positive correlation observed between GradSim and ripple effect performance across various LMs, KE methods, and evaluation metrics underscores its reliability.
Additionally, our exploration of failure cases further confirms that low GradSim values are indicative of ripple effect failures.

\section*{Limitations}
The first notable limitation is that, although a strong relationship between GradSim and the performance of ripple effects has been demonstrated, our research remains at the level of exploring correlations between these two factors and has not yet established a causal relationship. While it is always challenging to determine causality, it would still be extremely interesting and exciting to explore the dominant contributing factors to the complex distribution of knowledge storage in the pre-training phase of LMs.
The second important limitation is that, while this paper identifies an indicator, we did not provide practical solutions for improving ripple effect performance by leveraging this indicator. 
However, we believe that the insights provided in this short paper will significantly enable the development of practical and effective methods in future research.
% \section*{Ethics Statement}
% Scientific work published at EMNLP 2023 must comply with the \href{https://www.aclweb.org/portal/content/acl-code-ethics}{ACL Ethics Policy}. We encourage all authors to include an explicit ethics statement on the broader impact of the work, or other ethical considerations after the conclusion but before the references. The ethics statement will not count toward the page limit (8 pages for long, 4 pages for short papers).

\section* {Acknowledgement}
This research is based upon work supported by U.S. DARPA SemaFor Program No. HR001120C0123, DARPA Seedling BRIES HR0011-24-3-0325, DARPA INCAS Program No. HR001121C0165, and DARPA ITM Program No. FA8650-23-C-7316. The views and conclusions contained herein are those of the authors and should not be interpreted as necessarily representing the official policies, either expressed or implied, of DARPA, or the U.S. Government. The U.S. Government is authorized to reproduce and distribute reprints for governmental purposes notwithstanding any copyright annotation therein.

% \bibliographystyle{acl_natbib}
\bibliography{anthology,custom}

\begin{thebibliography}{14}
\providecommand{\natexlab}[1]{#1}

\bibitem[{AlKhamissi et~al.(2022)AlKhamissi, Li, Celikyilmaz, Diab, and Ghazvininejad}]{alkhamissi2022review}
Badr AlKhamissi, Millicent Li, Asli Celikyilmaz, Mona Diab, and Marjan Ghazvininejad. 2022.
\newblock A review on language models as knowledge bases.
\newblock \emph{arXiv preprint arXiv:2204.06031}.

\bibitem[{Cohen et~al.(2023)Cohen, Biran, Yoran, Globerson, and Geva}]{rippleedit}
Roi Cohen, Eden Biran, Ori Yoran, Amir Globerson, and Mor Geva. 2023.
\newblock \href {https://arxiv.org/abs/2307.12976} {Evaluating the ripple effects of knowledge editing in language models}.
\newblock \emph{Preprint}, arXiv:2307.12976.

\bibitem[{Liu et~al.(2024)Liu, Yu, Zhang, Li, Zhang, and Ji}]{liu2024evedit}
Jiateng Liu, Pengfei Yu, Yuji Zhang, Sha Li, Zixuan Zhang, and Heng Ji. 2024.
\newblock \href {https://arxiv.org/abs/2402.11324} {Evedit: Event-based knowledge editing with deductive editing boundaries}.
\newblock \emph{Preprint}, arXiv:2402.11324.

\bibitem[{Meng et~al.(2022)Meng, Bau, Andonian, and Belinkov}]{meng2022locating}
Kevin Meng, David Bau, Alex Andonian, and Yonatan Belinkov. 2022.
\newblock Locating and editing factual associations in gpt.
\newblock \emph{Advances in Neural Information Processing Systems}, 35:17359--17372.

\bibitem[{Meng et~al.(2023)Meng, Sharma, Andonian, Belinkov, and Bau}]{memit}
Kevin Meng, Arnab~Sen Sharma, Alex~J. Andonian, Yonatan Belinkov, and David Bau. 2023.
\newblock \href {https://openreview.net/pdf?id=MkbcAHIYgyS} {Mass-editing memory in a transformer}.
\newblock In \emph{The Eleventh International Conference on Learning Representations, {ICLR} 2023, Kigali, Rwanda, May 1-5, 2023}. OpenReview.net.

\bibitem[{Petroni et~al.(2019)Petroni, Rockt{\"a}schel, Riedel, Lewis, Bakhtin, Wu, and Miller}]{petroni2019language}
Fabio Petroni, Tim Rockt{\"a}schel, Sebastian Riedel, Patrick Lewis, Anton Bakhtin, Yuxiang Wu, and Alexander Miller. 2019.
\newblock Language models as knowledge bases?
\newblock In \emph{Proceedings of the 2019 Conference on Empirical Methods in Natural Language Processing and the 9th International Joint Conference on Natural Language Processing (EMNLP-IJCNLP)}, pages 2463--2473.

\bibitem[{Radford et~al.(2019)Radford, Wu, Child, Luan, Amodei, Sutskever et~al.}]{gpt2}
Alec Radford, Jeffrey Wu, Rewon Child, David Luan, Dario Amodei, Ilya Sutskever, et~al. 2019.
\newblock Language models are unsupervised multitask learners.
\newblock \emph{OpenAI blog}, 1(8):9.

\bibitem[{Song et~al.(2024)Song, Wang, He, Dong, Mou, Zhao, and Xu}]{song2024knowledge}
Xiaoshuai Song, Zhengyang Wang, Keqing He, Guanting Dong, Yutao Mou, Jinxu Zhao, and Weiran Xu. 2024.
\newblock \href {https://arxiv.org/abs/2402.08631} {Knowledge editing on black-box large language models}.
\newblock \emph{Preprint}, arXiv:2402.08631.

\bibitem[{Touvron et~al.(2023)Touvron, Martin, Stone, Albert, Almahairi, Babaei, Bashlykov, Batra, Bhargava, Bhosale, Bikel, Blecher, Ferrer, Chen, Cucurull, Esiobu, Fernandes, Fu, Fu, Fuller, Gao, Goswami, Goyal, Hartshorn, Hosseini, Hou, Inan, Kardas, Kerkez, Khabsa, Kloumann, Korenev, Koura, Lachaux, Lavril, Lee, Liskovich, Lu, Mao, Martinet, Mihaylov, Mishra, Molybog, Nie, Poulton, Reizenstein, Rungta, Saladi, Schelten, Silva, Smith, Subramanian, Tan, Tang, Taylor, Williams, Kuan, Xu, Yan, Zarov, Zhang, Fan, Kambadur, Narang, Rodriguez, Stojnic, Edunov, and Scialom}]{llama2}
Hugo Touvron, Louis Martin, Kevin Stone, Peter Albert, Amjad Almahairi, Yasmine Babaei, Nikolay Bashlykov, Soumya Batra, Prajjwal Bhargava, Shruti Bhosale, Dan Bikel, Lukas Blecher, Cristian~Canton Ferrer, Moya Chen, Guillem Cucurull, David Esiobu, Jude Fernandes, Jeremy Fu, Wenyin Fu, Brian Fuller, Cynthia Gao, Vedanuj Goswami, Naman Goyal, Anthony Hartshorn, Saghar Hosseini, Rui Hou, Hakan Inan, Marcin Kardas, Viktor Kerkez, Madian Khabsa, Isabel Kloumann, Artem Korenev, Punit~Singh Koura, Marie-Anne Lachaux, Thibaut Lavril, Jenya Lee, Diana Liskovich, Yinghai Lu, Yuning Mao, Xavier Martinet, Todor Mihaylov, Pushkar Mishra, Igor Molybog, Yixin Nie, Andrew Poulton, Jeremy Reizenstein, Rashi Rungta, Kalyan Saladi, Alan Schelten, Ruan Silva, Eric~Michael Smith, Ranjan Subramanian, Xiaoqing~Ellen Tan, Binh Tang, Ross Taylor, Adina Williams, Jian~Xiang Kuan, Puxin Xu, Zheng Yan, Iliyan Zarov, Yuchen Zhang, Angela Fan, Melanie Kambadur, Sharan Narang, Aurelien Rodriguez, Robert Stojnic, Sergey Edunov, and Thomas
  Scialom. 2023.
\newblock \href {https://arxiv.org/abs/2307.09288} {Llama 2: Open foundation and fine-tuned chat models}.
\newblock \emph{Preprint}, arXiv:2307.09288.

\bibitem[{Yang et~al.(2023)Yang, Xiao, Wang, Zhang, Bian, Yin, Lv, Pan, Wang, Yan, Yang, Deng, Wang, Liu, Ai, Dong, Zhao, Xu, Sun, Zhang, Liu, Ji, Xie, Dai, Fang, Su, Song, Liu, Ru, Ma, Wang, Liu, Lin, Nie, Guo, Sun, Zhang, Li, Li, Cheng, Chen, Zeng, Wang, Chen, Men, Yu, Pan, Shen, Wang, Li, Jiang, Gao, Zhang, Zhou, and Wu}]{yang2023baichuan}
Aiyuan Yang, Bin Xiao, Bingning Wang, Borong Zhang, Ce~Bian, Chao Yin, Chenxu Lv, Da~Pan, Dian Wang, Dong Yan, Fan Yang, Fei Deng, Feng Wang, Feng Liu, Guangwei Ai, Guosheng Dong, Haizhou Zhao, Hang Xu, Haoze Sun, Hongda Zhang, Hui Liu, Jiaming Ji, Jian Xie, JunTao Dai, Kun Fang, Lei Su, Liang Song, Lifeng Liu, Liyun Ru, Luyao Ma, Mang Wang, Mickel Liu, MingAn Lin, Nuolan Nie, Peidong Guo, Ruiyang Sun, Tao Zhang, Tianpeng Li, Tianyu Li, Wei Cheng, Weipeng Chen, Xiangrong Zeng, Xiaochuan Wang, Xiaoxi Chen, Xin Men, Xin Yu, Xuehai Pan, Yanjun Shen, Yiding Wang, Yiyu Li, Youxin Jiang, Yuchen Gao, Yupeng Zhang, Zenan Zhou, and Zhiying Wu. 2023.
\newblock \href {https://arxiv.org/abs/2309.10305} {Baichuan 2: Open large-scale language models}.
\newblock \emph{Preprint}, arXiv:2309.10305.

\bibitem[{Yin et~al.(2023)Yin, Jiang, Yang, and Wan}]{yin2023history}
Xunjian Yin, Jin Jiang, Liming Yang, and Xiaojun Wan. 2023.
\newblock \href {https://arxiv.org/abs/2312.05497} {History matters: Temporal knowledge editing in large language model}.
\newblock \emph{Preprint}, arXiv:2312.05497.

\bibitem[{Zhang et~al.(2024)Zhang, Li, Liu, Yu, Fung, Li, Li, and Ji}]{zhang2024knowledge}
Yuji Zhang, Sha Li, Jiateng Liu, Pengfei Yu, Yi~R Fung, Jing Li, Manling Li, and Heng Ji. 2024.
\newblock Knowledge overshadowing causes amalgamated hallucination in large language models.
\newblock \emph{arXiv preprint arXiv:2407.08039}.

\bibitem[{Zhong et~al.(2023{\natexlab{a}})Zhong, Wu, Manning, Potts, and Chen}]{zhong-etal-2023-mquake}
Zexuan Zhong, Zhengxuan Wu, Christopher Manning, Christopher Potts, and Danqi Chen. 2023{\natexlab{a}}.
\newblock \href {https://doi.org/10.18653/v1/2023.emnlp-main.971} {{MQ}u{AKE}: Assessing knowledge editing in language models via multi-hop questions}.
\newblock In \emph{Proceedings of the 2023 Conference on Empirical Methods in Natural Language Processing}, pages 15686--15702, Singapore. Association for Computational Linguistics.

\bibitem[{Zhong et~al.(2023{\natexlab{b}})Zhong, Wu, Manning, Potts, and Chen}]{zhong2023mquake}
Zexuan Zhong, Zhengxuan Wu, Christopher~D Manning, Christopher Potts, and Danqi Chen. 2023{\natexlab{b}}.
\newblock Mquake: Assessing knowledge editing in language models via multi-hop questions.
\newblock \emph{arXiv preprint arXiv:2305.14795}.

\end{thebibliography}

\newpage

\begin{figure*}[htbp]
  \centering
  \includegraphics[width=\textwidth]{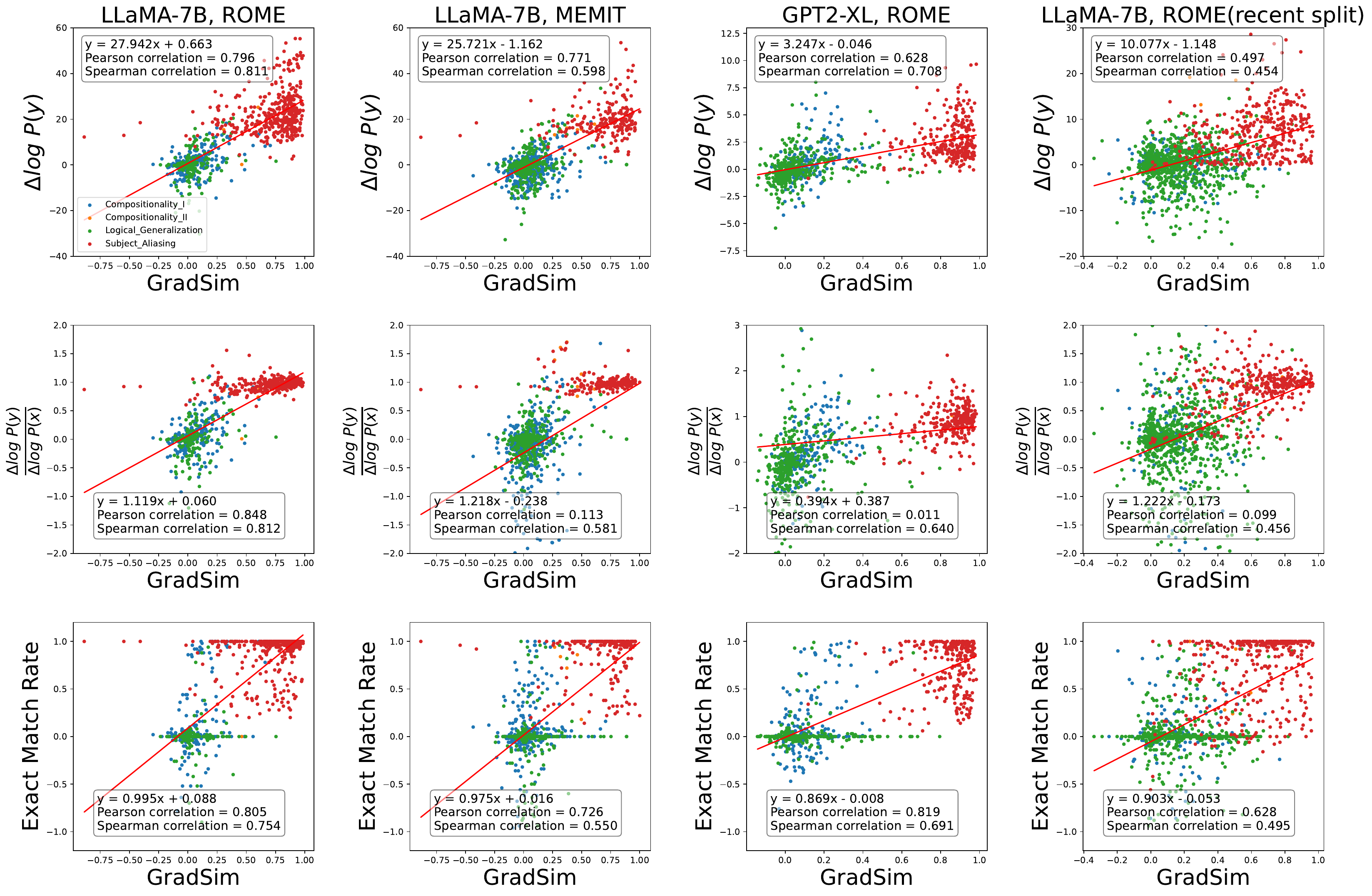} 
  \caption{Main results of evaluating the correlation between ripple effect performances and GradSim values labeled with different task names}
  \vspace{-0.4cm}
  \label{fig:wideimage2}
\end{figure*}

\appendix
\section{Knoweledge Editing Methods}

Current knowledge editing methods can be categorized into several distinct approaches, including fine-tuning-based methods, locate-then-edit methods, and meta-learning methods. In this paper, we employ a range of knowledge editing techniques to validate our theorem. A brief introduction to each of these methods will be provided here.

% \textbf{Fine-Tuning}: 
\textbf{ROME}: Rank-One Model Editing (ROME) conceptualizes an MLP module as a key-value store. In this framework, the key represents an encoded subject, while the value represents the knowledge associated with that subject. The MLP retrieves the corresponding value by accessing the key. ROME modifies the MLP weights using a rank-one adjustment to directly insert new key-value pairs.

\textbf{MEMIT}: Mass Editing Memory in a Transformer (MEMIT) builds upon ROME, and is designed to handle large-scale edits by inserting multiple memory entries simultaneously through modifications to the MLP weights across several key layers. MEMIT employs causal tracing to identify a set of mediating MLP layers that store and recall memories related to a specific subject. For a set of new memories, an update ($\Delta$) is computed and propagated across all the identified mediating MLP layers, ensuring that by the final layer, the output captures and reflects all the newly inserted memories.

% \textbf{Fine-tuning}: 

\textbf{MEND}: Model Editor Networks with Gradient Decomposition (MEND) consist of small auxiliary networks designed to make quick, localized edits to a pre-trained model’s behavior using a single input-output pair. MEND achieves this by learning how to transform the gradient generated through standard fine-tuning, employing a low-rank decomposition of the gradient to make the transformation more computationally feasible. MEND can be trained efficiently on a single GPU in less than a day, even for models with over 10 billion parameters. Once trained, MEND allows for rapid application of new edits to the pre-trained model.

\section{Additional Experiments}

\subsection{Experiments with various knowledge experiments}
To support our theorem, we also did experiments with more knowledge editing methods, including Fine-Tuning(FT) and MEND. In the basic FT process, we use Adam and early stopping to minimize the loss of new edits on full parameters.  In the additional experiments, we only compute on GPT-XL due to computational resource limitations. The experiments are done with the popular subset in RippleEdits.
Here are the experiment results:

\begin{table}[h!]
\centering
\resizebox{\columnwidth}{!}{%
\begin{tabular}{c|c|c|c}
\hline
\textbf{Method}  & \textbf{Spearman Correlation} & \textbf{Pearson Correlation} & \textbf{Best Fit Line} \\ 
\hline
MEND &  0.628 & 0.501 & $y = 0.845x - 0.177$ \\
\hline
FT & 0.231 & 0.731 & $y = 0.430x + 0.119$ \\
\hline
\end{tabular}%
}
\caption{Main experiments with other knowledge editing method}
\end{table}
These experimental results demonstrate the versatility of the GradSim indicator.

\subsection{Experiments on other dataset}
We conducted experiments on all data in RippleEdits to demonstrate the indicator's effectiveness. Additional experiments are also done on the 2-hop subset of MQUAKE dataset~\cite{zhong2023mquake} to support the validity of our result. The results are as follows

\begin{table}[h!]
\centering
\resizebox{\columnwidth}{!}{%
\begin{tabular}{c|c|c|c|c}
\hline
\textbf{Method} & \textbf{Model} & \textbf{Spearman Correlation} & \textbf{Pearson Correlation} & \textbf{Best Fit Line} \\ 
\hline
ROME & GPT2-XL & 0.767 & 0.783 & $y = 1.164x - 8.923$ \\
\hline
\end{tabular}%
}
\caption{Results on MQUAKE}
\end{table}

\section{Extra Results of GradSim}
The RippleEdits dataset comprises six distinct tasks: Logical Generalization (LG), Compositionality I (CI), Compositionality II (CII), Subject Aliasing (SA), Preservation (PV), and Relation Specificity (RS)~\citep{rippleedit}. These tasks are designed to evaluate different aspects of knowledge editing in neural networks. Specifically, LG, CI, CII, and SA are tasks where the model is expected to manifest new knowledge in response to edits made to existing entries. Conversely, in the RS and PV tasks, the existing knowledge should remain unaltered post-editing, as these tasks are designed to test the model's ability to preserve information that is logically independent of the changes applied.

\begin{figure}[htbp]
  \centering
  \includegraphics[width=\linewidth]{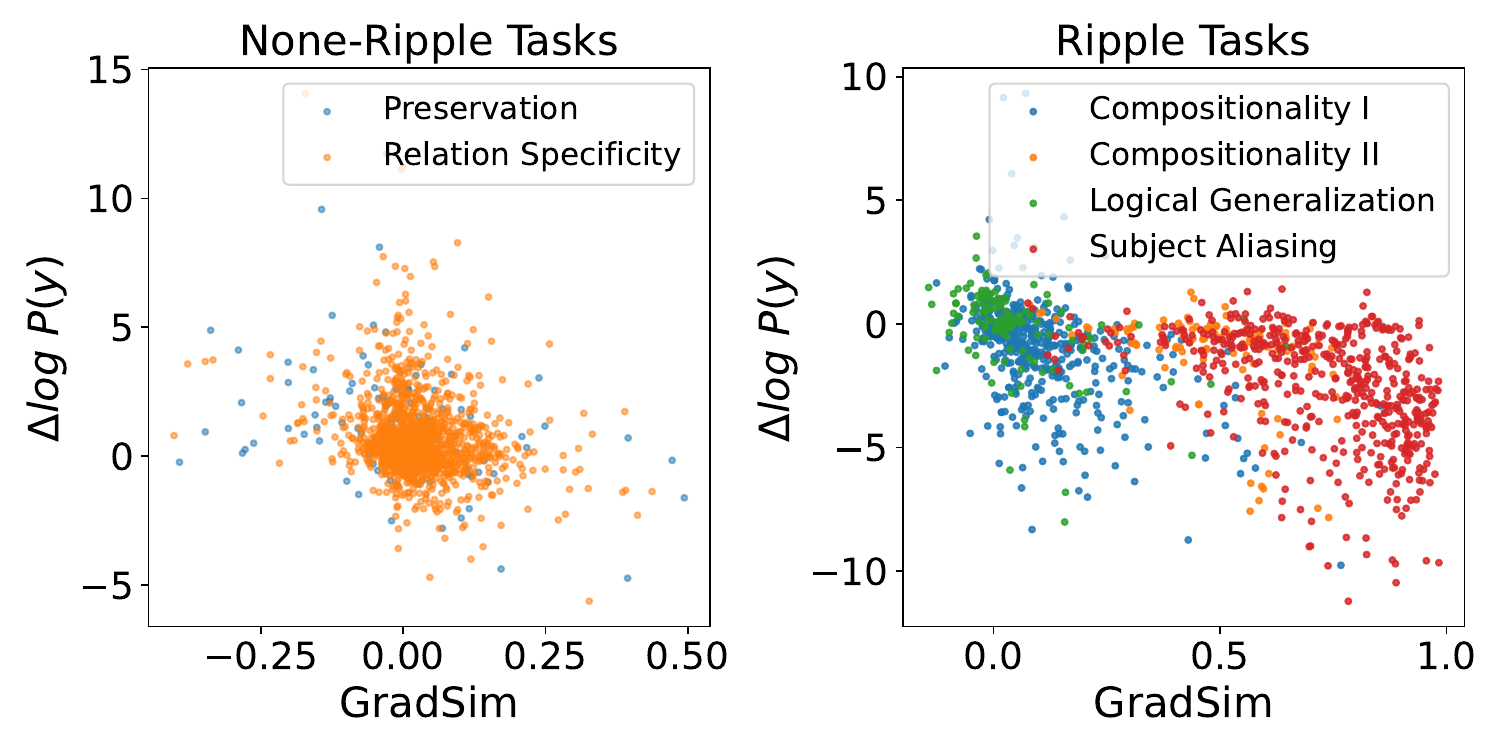}
  \caption{Comparison of the Correlation on None-Ripple Tasks and Ripple Tasks.}
  % \vspace{-0.55cm}
  \label{fig:comparision}
\end{figure}

In Figure \ref{fig:wideimage}, we analyze data from these tasks to illustrate a positive correlation between ripple effect performance and GradSim values across the four tasks explicitly associated with knowledge updates, hereafter referred to as \emph{ripple tasks}. Each data point in Figure \ref{fig:wideimage2} is labeled to demonstrate the consistent applicability of the GradSim metric across the individual sub-tasks. In contrast, Figure \ref{fig:comparision} focuses on the two \emph{non-ripple tasks} (RS and PV), where no significant correlation is observed between GradSim values and $\Delta \log P(y)$. The comparison underscores that GradSim is a critical metric for evaluating ripple effects, as it shows no significant impact in the \emph{non-ripple tasks}, confirming its relevance specifically in contexts where knowledge modifications are expected.

\section{How to Represent Knowledge Distribution?}

In this paper, we utilize gradients to represent the distribution of knowledge. To support this approach, we conducted preliminary experiments that lend credence to the underlying rationale of this intuition.

\subsection{Does the way that we express a piece of knowledge change the knowledge distribution?}
\begin{figure}[htbp]
  \centering
  \includegraphics[width=\linewidth]{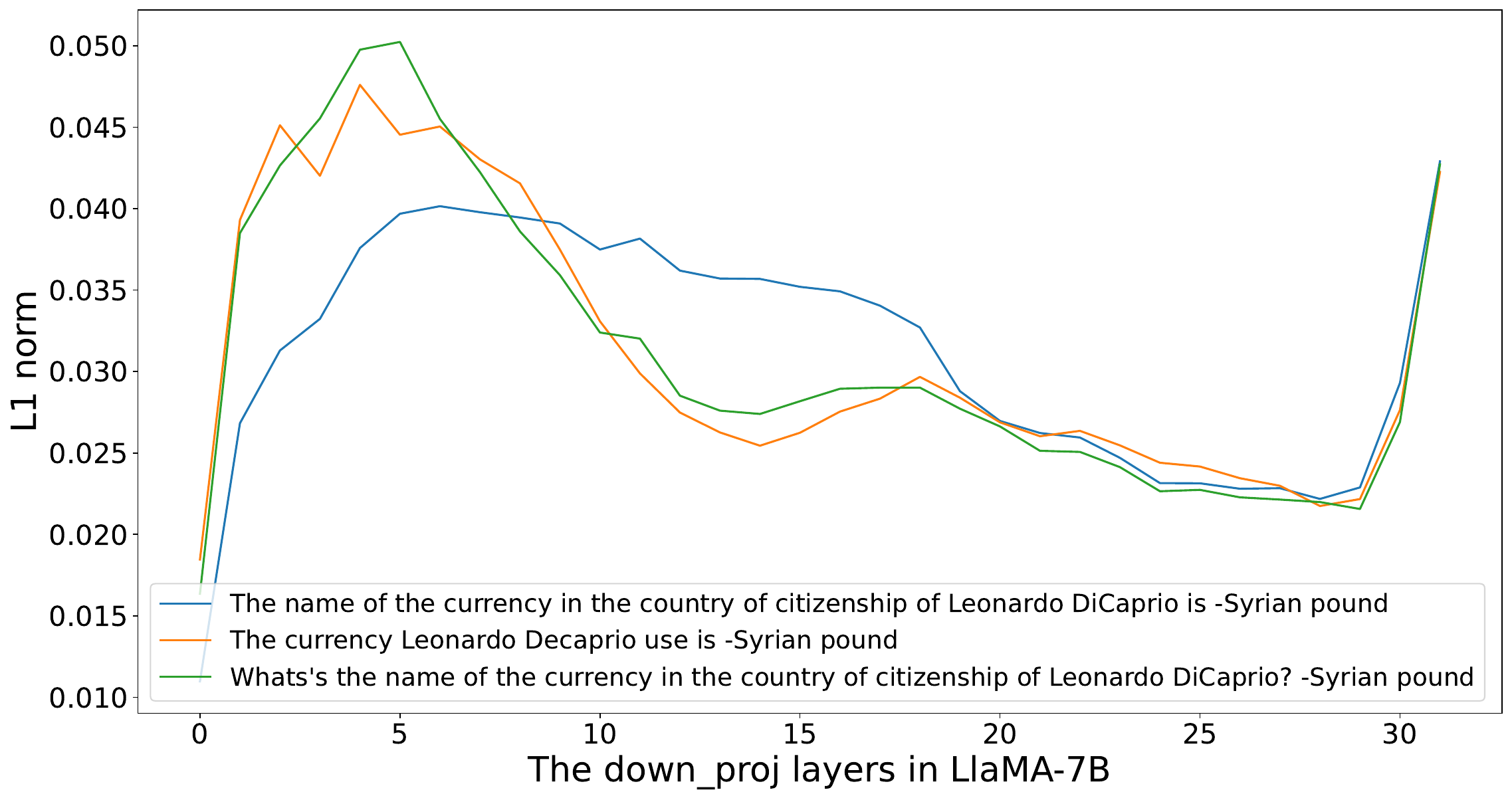}
  \caption{L1 Norm Distribution over LlaMA-7B}
  % \vspace{-0.55cm}
  \label{fig:l1_norm}
\end{figure}
In this study, we calculate the gradient of a specific piece of knowledge, "The name of the currency in the country of citizenship of Leonardo DiCaprio is the Syrian pound," along with its variants: "The currency Leonardo DiCaprio uses is the Syrian pound" and "What's the name of the currency in the country of citizenship of Leonardo DiCaprio? Syrian pound." We then plot the L1 norm of the gradient across the 32 downward projection layers of LlaMA7b. Prior research suggests that these layers are particularly adept at storing knowledge. Our results indicate that the distribution across these variants is remarkably consistent, suggesting that a piece of knowledge may be encoded within specific parameters of a large language model.

\label{sec:appendix}

\end{document}